\documentclass[11pt]{article}

\usepackage[final]{acl}

\usepackage{times}
\usepackage{latexsym}

\usepackage[T1]{fontenc}

\usepackage[utf8]{inputenc}

\usepackage{microtype}

\usepackage{inconsolata}

\usepackage{graphicx}

\usepackage{multirow}
\usepackage{amsmath}
\usepackage{makecell}
\usepackage{amssymb}
\usepackage{soul}
\usepackage{xcolor}
\sethlcolor{yellow}

\title{Applicability Condition Extraction for Therapeutic Drug-Disease Relations}

\author{
  Guanting Luo$^{1,2}$ \quad Noriki Nishida$^{2}$ \quad Yuji Matsumoto$^{2,4}$ \quad Yuki Arase$^{3,2}$ \\
  $^{1}$The University of Osaka \quad $^{2}$RIKEN \quad $^{3}$Institute of Science Tokyo \quad $^{4}$Tohoku University\\
  \texttt{guanting.luo@ist.osaka-u.ac.jp} \\
  \texttt{noriki@norikinishida.com} \\
  \texttt{yuji.matsumoto@riken.jp} \\
  \texttt{arase@c.titech.ac.jp}
}

\begin{document}
\maketitle
\begin{abstract}
Identifying conditions that a certain drug takes therapeutic effect on a target disease is crucial for clinical decision-making support.
However, most existing biomedical information extraction methods have focused on identifying only relations between drugs and diseases,
while largely overlooking the context-specific conditions where such relations can apply.
To address this problem, we introduce the task of applicability condition extraction for therapeutic drug--disease relations from biomedical research literature.
We create the first dataset that has manually annotated triples of drugs, diseases, and applicability conditions on biomedical paper abstracts with $1,119$ drug-disease pairs.
Using this dataset, we systematically evaluate the performance of a range of existing methods.
In addition, we propose a new method that enhances LoRA to consider relations between drugs and diseases. Our method consistently outperforms strong baselines across different evaluation settings.
The source code and dataset of this paper can be obtained from: \url{https://github.com/guantingluo98/Drug-ACE}
\end{abstract}

\section{Introduction}

Therapeutic drug--disease relations play a central role in clinical practice and biomedical research, forming the foundation for treatment selection and evidence-based medical decision-making.
In real-world clinical settings, however, the applicability of a drug is rarely universal across all patient populations.
Whether a drug can be effective to treat a disease often depends on specific patient profiles and contextual factors, reflecting substantial patient heterogeneity.
Therefore, it is crucial to identify conditions under which a drug can be effectively and safely applied to treat a target disease. 
These applicability conditions are critical for translating biomedical evidence into practical clinical decision-making and for accurately interpreting therapeutic claims reported in the biomedical research literature. 
However, such applicability condition extraction has been little explored despite abundant research efforts on drug-disease relation extraction~\citep{wei2016assessing, nguyen-verspoor-2018-convolutional, bonner2022review, luo2022biored, wang-etal-2024-bio}. 

\begin{figure}[t]
\centering
\small
\begin{tabular}{p{0.95\columnwidth}}
\hline
\textbf{Example of an Annotated Instance} \\
\hline
\textbf{Title}: Hydroxyurea in stage D carcinoma of the prostate: a pilot study. \\

\textbf{Abstract}: There was 13 patients with histologically metastatic \textbf{prostatic adenocarcinoma} treated with \colorbox{yellow!70}{a single oral dose of 80 mg. per kg. \textbf{hydroxyurea} every} \colorbox{yellow!70}{third day} (based on ideal or actual weight, whichever is less) and 12.5 mg. chlorotrianisene per day. Toxicity was mild. The most common manifestations were nausea, occasional vomiting leukopenia. A definite attempt was made to depress the white blood count to approximately 2,000 cells per cu. mm. Hydroxyurea was not discontinued unless the white blood count decreased to less than 2,000 cells per cu. mm., after which a single dose was usually omitted. Omission of a single dose would allow the white blood count to return promptly to more than 2,000 cells per cu. mm. Objective tumor regression was demonstrated in 6 of the 13 patients and all patients had a definite improvement in the quality of life. \\
\hline
\textbf{Drug--Disease Pair}: (Hydroxyurea, prostatic adenocarcinoma) \\
\textbf{Applicable Condition}: ``a single oral dose of 80 mg. per kg. hydroxyurea every third day'' \\ 
\textbf{Condition Type}: Dosage \\
\hline
\end{tabular}
\caption{An illustrative example of an annotated instance in the Drug-ACE dataset.}
\label{tab:example-instance}
\end{figure}

In biomedical research literature, therapeutic evidence is frequently reported in a conditional manner rather than as universally applicable conclusions \citep{lu2011pubmed}.
The effectiveness of a drug is often qualified by specific conditions, such as dosage, patient populations, physiological characteristics, comorbidities, or genetic background. These conditions reflect the inherent diversity of patients
and the complexity of disease mechanisms,
and are essential for accurately interpreting therapeutic claims \citep{weinshilboum2017pharmacogenomics}.
Unfortunately, such applicability conditions are rarely stated in an explicit manner in a document (Figure~\ref{tab:example-instance}).
Instead, they are often distributed across sentences,
embedded within broader experimental or clinical descriptions, and expressed implicitly through contextualized evidence.
As a result, understanding therapeutic applicability requires reasoning over lengthy and nuanced textual contexts.

Despite substantial progress in biomedical information extraction, most existing studies have primarily focused on identifying whether a certain relation exists between biomedical entities \citep{roy-pan-2021-incorporating, jin-etal-2022-supporting, xiao-etal-2024-federated}, or on extracting specific phenomena such as adverse effects \citep{alimova-tutubalina-2019-detecting, henry20202018, doosterlinck-etal-2023-biodex, sahoo-etal-2024-enhancing}.

However, the conditions under which the relation is applicable or not have been little explored. 
As a result, current biomedical information extraction frameworks often provide incomplete representations of therapeutic knowledge, limiting their usefulness for clinical decision support.

To address this problem, we extract conditions for therapeutic drug--disease relations. 
Specifically, we create the Drug--Disease Applicability Condition Extraction (Drug-ACE) dataset as illustated in Figure~\ref{tab:example-instance}.

The dataset contains $1,119$ instances, each associated with a therapeutic drug--disease pair and corresponding PubMed paper titles and abstracts.
Each instance is annotated with the conditions under which the given drug can treat or alleviate the target disease.

We also propose a Role-Conditioned LoRA that explicitly incorporates the relation role between the drug and disease into parameter-efficient low-rank adaptation (LoRA) \citep{hu2022lora}. 
Our comprehensive benchmarking study on the Drug-ACE dataset comparing existing biomedical relation extraction methods reveals that our method consistently outperforms strong baselines.

Our contributions are threefold:
\begin{itemize}
  \item We introduce the task of drug--disease applicability condition extraction and release Drug-ACE, annotating $1,119$ instances.
  \item We deliver a comprehensive evaluation of conventional biomedical relation extraction methods on Drug-ACE, including span-based models, LoRA-tuning and prompting large language models.
  \item We propose a method for applicable condition extraction of drug-disease relations, which consistently outperforms strong baselines across different evaluation settings.
\end{itemize}

\section{Related Work}

\subsection{Biomedical Relation Extraction}

Biomedical relation extraction has been extensively studied as a core task in biomedical natural language processing,
with particular attention to identifying relations between chemicals, diseases, and genes from scientific literature.
Early benchmark efforts, such as the BioCreative V CDR \citep{li2016biocreative} task corpus,
established standard evaluation settings for chemical–disease relation extraction and facilitated the development of supervised learning approaches.
Subsequent work further expanded the scope of biomedical relation extraction by constructing larger and richer datasets, including BioRED \citep{luo2022biored} and ChemDisGene~\citep{zhang-etal-2022-distant}, which cover diverse entity types and multiple relation categories. DrugProt \citep{miranda2023overview} introduced a large-scale gold standard for granular drug-gene/protein interactions.
\citet{SOSA2023104474} frame the association of cell types and tissues with protein-protein interactions as a classification task, utilizing syntactic and meta-discourse features to enrich literature-derived knowledge graphs.

In parallel, researchers have explored alternative learning paradigms
to address data sparsity and annotation cost in biomedical relation extraction.
\citet{xiao-etal-2024-federated} extend document-level relation extraction
to a federated learning setting for the first time
and propose a novel non--independently and identically distributed scenario based on graph structural entropy.
\citet{wang-etal-2024-bio} investigate a pipeline that performs sentence-level relation classification prior to entity extraction to alleviate entity ambiguity,
and further incorporate structural constraints between entities and relations to guide the model’s hypothesis space.

These studies have significantly advanced the modeling of biomedical relations under various practical constraints.
However, despite these efforts, existing work has primarily focused on identifying
the presence or type of relations between entities,
and does not explicitly model the applicability conditions under which therapeutic drug--disease relations hold.

\subsection{Fine-grained Information Extraction Benchmarks}

Beyond relation extraction, prior work has explored a variety of fine-grained information extraction tasks
that aim to identify condition-like or attribute-level information from textual data.
In the clinical domain, adverse event extraction has been studied as a representative task,
where models are required to extract specific event spans and assign them to predefined categories 
\citep{doosterlinck-etal-2023-biodex, sahoo-etal-2024-enhancing, guellil-etal-2025-adverse}.
Early approaches typically rely on a sequence tagging framework, such as conditional random fields (CRFs),
to model token-level dependencies and capture structured output constraints ~\citep{guellil-etal-2025-adverse}. 
\citet {srivastava-etal-2025-instruction}
explore instruction-tuning on large language models (LLMs) for event extraction, leveraging textual annotation guideline to guide model predictions.
Their results show that prompt- and instruction-based approaches can serve as effective alternatives to traditional supervised models, particularly in low-resource or cross-schema settings.
These findings support the use of prompting-based methods as a reasonable approach for fine-grained information extraction tasks.

\section{Drug-ACE Dataset}
We create the Drug-ACE dataset that manually annotates conditions under which a certain drug takes a therapeutic effect against a disease. 
Figure~\ref{tab:example-instance} presents an example, showing the input and the annotated applicability condition. 

\subsection{Annotation Data Preparation}

Our dataset is constructed on top of the ChemDisGene dataset~\citep{zhang-etal-2022-distant},
which consists of PubMed\footnote{\url{https://pubmed.ncbi.nlm.nih.gov/}} biomedical abstracts annotated with drug, disease, and gene entities,
as well as pairwise relations among them.
 
The original ChemDisGene dataset covers a diverse set of entity types and relation categories.
In this work, we restrict our focus to instances involving \textit{therapeutic} drug--disease relations for their practical values 
and filter out instances corresponding to other relation types.
The original ChemDisGene includes relations identified by \textit{in vivo} and \textit{in vitro} experiments. 
We manually reviewed and further filtered those that do not mention clinical studies or clinical trials,
retaining only clinically grounded drug--disease relations for applicability condition annotation.
We further manually inspected the therapeutic relation annotations and removed instances which seem to have incorrect or inconsistent 
relation annotations.

Note that our preprocessing step does not modify the original textual content of the biomedical literature. 
Rather, we only restrict instances included in Drug-ACE to be ones with therapeutic relations with sufficiently reliable evidence.

\subsection{Condition Types}

To better understand applicability conditions and facilitate their systematic modeling, we assign type labels to each extracted condition. 
There has been no consensus on an exhaustive taxonomy of applicability conditions for drug--disease relations.
To design a reasonable set of condition types, we reviewed relevant biomedical literature \citep{wu2019renet, bhatt2021dice, hanlon2023treatment} and identified commonly discussed conditions
that constrain or qualify therapeutic applicability. 
We then consult with a domain expert 
to define the annotation scope, and finally selected six condition types after multiple rounds of discussions.

The following six condition types are frequently observed in biomedical literature and are clinically meaningful which can substantially influence treatment effectiveness and clinical decision-making. 
\begin{itemize}
  \item \textit{Dosage}, indicates the dosage or amount of the drug administered, including specific dosage values and ranges required for effective or safe treatment. E.g.: ``\textbf{3.3 mg/70 kg of M6G}''
  \item \textit{Age}, specifies the patient's age or age group, including explicit ages or age-related categories that affects the applicability of the treatment. E.g.: ``\textbf{children} with nephrotic syndrome'' 
  \item \textit{Gene}, specifies genetic characteristics of patients, such as the presence of particular genes, that influence drug response or treatment suitability. E.g.: ``\textbf{HDL-bound paraoxonase-1 (PON1)}'' 
  \item \textit{Gender}, indicates the biological sex or gender of the target patient population when treatment applicability differs across genders. E.g.: ``Fourteen \textbf{male} patients'' 
  \item \textit{Comorbidity}, refers to the presence of one or more additional pre-existing diseases,
  disorders, or risk factors other than the index disease under investigation, which may affect the drug’s applicability or therapeutic effect. E.g.: ``children with \textbf{nephrotic syndrome}''
  \item \textit{Body Type}, describes general body characteristics or physical conditions of patients,
  including pregnancy, obesity, underweight status, or other body composition–related factors that may influence treatment outcomes. E.g.: ``\textbf{primigravida} woman''
\end{itemize}

\subsection{Annotation}
The annotation has been conducted from Oct. to Dec., 2025. 
The primary communication tool was Slack; the annotators could ask questions anytime. 

\paragraph{Annotator Selection}
We recruited two graduate students majoring in life science and technology, and pathological biochemistry, respectively. 
To ensure annotation quality, these annotators were first asked to conduct trial sessions, during which their domain knowledge and understanding of the annotation guideline were carefully evaluated. 
The annotators were paid about $\$1$ per instance. 

\paragraph{Annotation Guideline}
We provided annotation guidelines to the annotators, which consists of task definition, inclusion and exclusion criteria, taxonomy for the condition types, and annotation scope.
The complete version of the guideline is presented in the Appendix~\ref{sec:guideline}.

\paragraph{Annotation Procedure}
We provided $200$ abstracts as a batch to the annotators to control the annotation quality. 
Each abstract was independently annotated by the two annotators. 
To ensure the consistency and reliability of the annotated applicability conditions, one of the authors reviewed all the results. 
In cases of disagreement, the third annotator (one of the authors) served as a judge
and determined the final annotation through adjudication. 
The initial agreement between the two annotators was $61\%$. 
The revised annotations were feedback to the annotators to improve the task understanding and agreement in the next batch. The agreement rate of the final batch improved to $86\%$. 

\subsection{Resultant Dataset}
\begin{figure}[t]
  \centering
  \includegraphics[width=\linewidth]{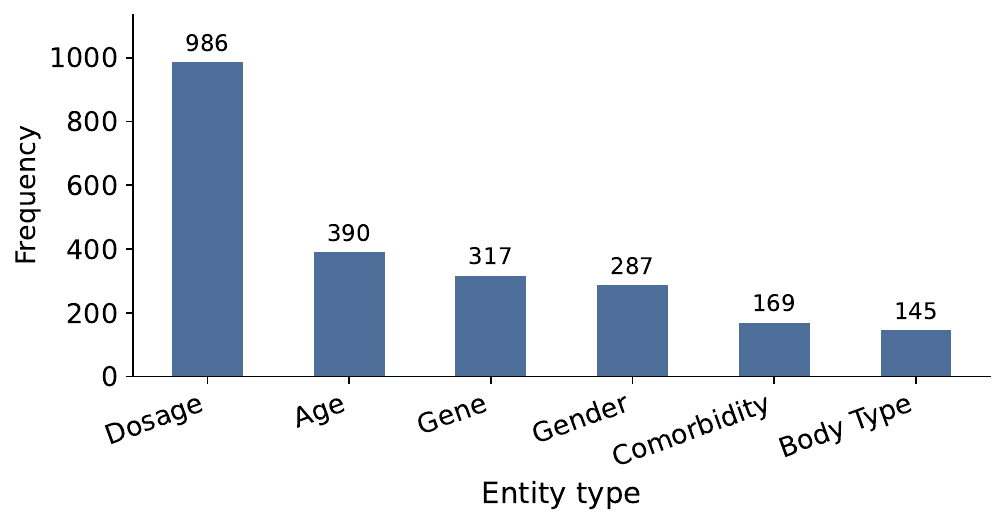}
  \caption{The distribution of applicability conditions.}
  \label{fig:type}
\end{figure}

As the final outcome of our annotation, our Durg-ACE dataset in total consists of $1,119$ drug-disease pairs from $667$ unique Pubmed abstracts associated with conditions. 
Namely, $2,290$ applicability condition spans were identified, with an average of approximately two annotated spans per drug-disease pair. 
Figure~\ref{fig:type} shows the distribution of condition types. The majority type was Dosage, followed by Age, Gene, and Gender. 

We split the dataset for training, development, and testing. 
Table~\ref{tab:data-stats} summarizes the dataset statistics for each split: the number of unique Pubmed abstracts (\textit{\#Abstracts}), the number of drug-disease pairs (\textit{\#Pairs}), the average document length measured in tokens (\textit{DocLen}), and the average number of applicability condition spans per pair (\textit{\#Spans}).

\begin{table}[t]
\centering
\small
\begin{tabular}{l c c c c}
\hline
\textbf{Category} & \textbf{\#Abstracts} & \textbf{\#Pairs}  & \textbf{DocLen} & \textbf{\#Spans} \\
\hline
Train & $334$  & $558$  & $266.0$ & $2.01$ \\
Dev   & $110$  & $182$  & $270.3$ & $2.38$ \\
Test  & $223$ & $379$  & $283.4$ & $1.94$ \\
\hline
\end{tabular}
\caption{Dataset statistics for the train, development, and test categories.}
\label{tab:data-stats}
\end{table}

\section{Applicability Condition Extraction}

We propose a method to extract applicability conditions of a given therapeutic drug-disease pair from biomedical paper titles and abstracts. 
Our method utilizes LLMs' strong capability in understanding long-texts and conduct parameter-efficient tuning with LoRA to predict conditions. 
To address the complex nature of biomedical literature understanding, we propose to explicitly model \emph{relation roles} between the drug and disease. 
Following \citet{he-etal-2025-tablelora} who employ tabular structure information for table understanding, we explicitly encode the relation roles in LoRA.  

\subsection{Task Definition}
We first define the task of therapeutic drug--disease applicability condition extraction. 
Given a pair of drug and disease, the goal is to identify conditions in texts under which the drug takes a therapeutic effect to treat the disease. More specifically, the input and output consist of the following.  

\paragraph{Input}
An input consists of (i) a therapeutic drug--disease pair and
(ii) the title and abstract of a biomedical paper where the pair is mentioned.
The title and abstract provide broader biomedical context, from which the applicability conditions for the given drug--disease pair can be extracted.

\paragraph{Output}
The output is a set of applicability conditions of the given drug--disease pair.
Each condition specifies a particular circumstance under which the drug is applicable, and is associated with predefined condition types.

\subsection{Relation Roles}
Drug–disease applicability condition extraction requires identifying conditions distributed across sentences from broader experimental or clinical descriptions, which can be implicitly mentioned. 
Therefore, it is insufficient to look for possible spans appearing close to the drug-disease pair. 
Furthermore, multiple drugs and diseases may co-occur, thus multiple conditions co-exist in text. 
Another hurdle is that biomedical entities tend to be split into subword tokens, which dilutes span-level representations~\citep{DBLP:conf/ijcai/BaldeRMG24}.
These unique characteristics easily confuse a model from understanding ``who applies to whom'' from context. 

To address these challenges, we encode the subject and object of a target condition into LoRA. 
Given the input consisting of a biomedical research literature $T=\{t_i\}_{i=1}^{|T|}$ (title and abstract) and a queried drug--disease pair $(d, s)$, we assign a relation role to each input token to explicitly encode its participation in the queried relation. 
Specifically, each token is labeled as one of three roles: \textsc{Obj} for tokens belonging to the given drug $d=\{d_1, d_2,\ldots,d_n\}$, \textsc{Subj} for tokens belonging to the given disease $s=\{s_1, s_2, \ldots, s_m\}$, and \textsc{NA} for all remaining tokens. 

Formally, we determine a role sequence $\{y_i\}_{i=1}^{|T|}$:
\begin{equation}
y_i =
\begin{cases}
\textsc{Obj}, & t_i \in d, \\
\textsc{Subj}, & t_i \in s, \\
\textsc{NA}, & \text{otherwise}.
\end{cases}
\end{equation}
We first locate the character-level spans of $d$ and $s$ in $T$ by exact matching after lemmatization, and then map these spans to token indices using tokenizer offsets. 
When an abstract contains multiple mentions of the same drug and disease string, we assign the same role to all of their occurrences. 
This role labelling is model-agnostic and introduces only minimal overhead while providing an explicit subject--object signal for subsequent LoRA tuning.

\subsection{Role-Conditioned LoRA}
\paragraph{Preliminary: LoRA}
We briefly review standard LoRA~\citep{hu2022lora} before introducing our method.
Given a frozen weight matrix $W_0 \in \mathbb{R}^{d \times k}$ of an LLM, LoRA models the parameter update $\Delta W$ as a low-rank decomposition.
\begin{equation}
\label{eq:standard lora}
 \mathbf{h}={W}_{0} \mathbf{x} + \Delta W \mathbf{x} = {W}_{0} \mathbf{x} + B A \mathbf{x},
\end{equation}
where $B \in \mathbb{R}^{d \times r}$ and $A \in \mathbb{R}^{r \times k}$. 
By assuming that the parameter update is low rank, i.e., $r \ll \min(d, k)$, LoRA significantly reduces the parameters to tune. 

\paragraph{Proposed Method: Role-Conditioned LoRA}
In the original LoRA, the low-rank update is applied uniformly across all tokens, without distinguishing the relation roles of different spans in the input, which can result in suboptimal performance \cite{zhao-etal-2023-infusing, pu-demberg-2024-rst}.

\begin{figure}[t]
  \includegraphics[width=\linewidth]{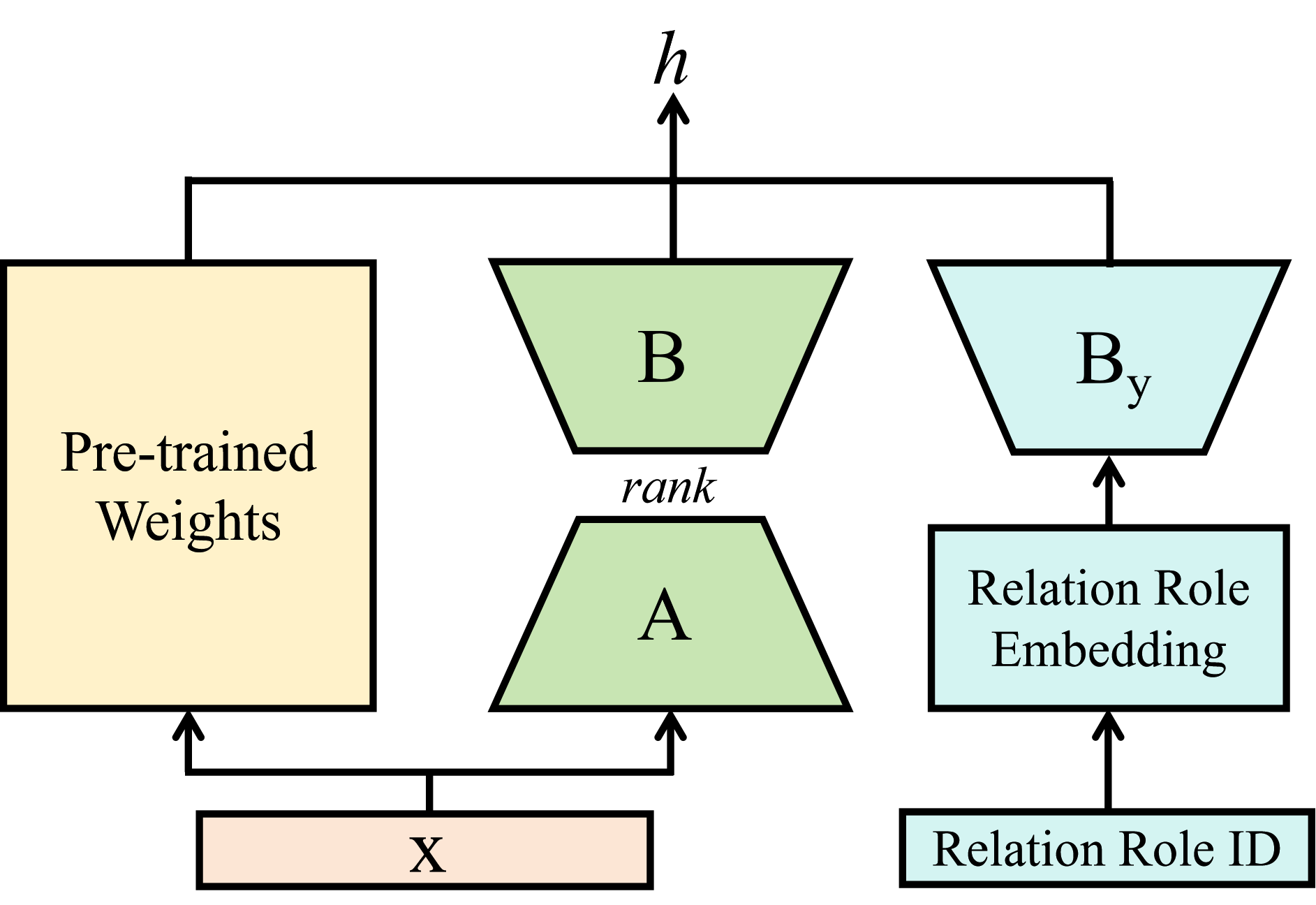}
  \caption {Model architecture: The diagram illustrates the integration of the relation role into the LoRA model. The left side is the standard LoRA, while the right side depicts our method.}
  \label{fig:method}
\end{figure}

We guide the fine-tuning process by injecting the relation role into LoRA, as shown in Figure~\ref{fig:method}. 
Specifically:
\begin{equation}
\label{eq:rc lora}
 \mathbf{h}={W}_{0} \mathbf{x} + B A \mathbf{x} + B_y\, \mathbf{e}_{y},
\end{equation}
where $y \in \{\textsc{Obj}, \textsc{Subj}, \textsc{NA}\}$ denotes the relation role assigned to the input token, $\mathbf{e}_{y}  \in \mathbb{R}^r$ is a learnable embedding corresponding to the role $y$, and $B_y \in \mathbb{R}^{d \times r}$ is a role-conditioned low-rank projection matrix.
The matrix $B_y$ shares the same shape as the original LoRA matrix $B$, and is initialized to zero in the same manner.

Note that the role embedding $\mathbf{e}_{y}$ is unique for each transformer layer.
This design allows different layers to capture role information at varying levels of abstraction.
This allows greater flexibility and expressiveness compared to sharing the same embedding across layers, while incurring only minimal additional parameters.

Finally, the LLM outputs a list of extracted applicability conditions,
with each item consisting of a textual span and an associated type, formatted as \texttt{Span: <span> | Label: <type>}. 
We employed the standard cross-entropy loss for training. 

\section{Experiment Setup}
Our Drug-ACE is the first dataset for drug–disease applicability condition extraction, thus we benchmark strong baseline methods on this dataset to clarify challenges in applicability condition extraction.
We also empirically evaluate the effectiveness of the proposed method. 

\subsection{Baselines}
We evaluate the following existing methods\footnote{We also experimented with token-level BERT-BiLSTM-CRF and BERT-CRF baselines \citep{hochreiter1997long,lafferty2001conditional,akbik-etal-2019-flair}. Despite careful tuning under the same training and evaluation protocols, these models ended up predicting no condition span, likely due to insufficient training samples. We therefore omit these baselines from the main comparison.}:
\begin{itemize}
\item \textbf{SpanMarker}\footnote{\url{https://tomaarsen.github.io/SpanMarkerNER/}} adhering to the Packed Levitated Marker architecture \citep{ye-etal-2022-packed}. Each model leverages a pre-trained language model as the backbone encoder, including RoBERTa (base and large) \citep{liu2019robertarobustlyoptimizedbert}, BERT (cased and uncased) \citep{devlin-etal-2019-bert}, BiomedBERT \citep{gu2021domain}, BioBERT \citep{lee2020biobert}, and Bio\_ClinicalBERT \citep{alsentzer-etal-2019-publicly}. Input sequences are processed using a levitated marker mechanism, in which pairs of trainable marker tokens are inserted into the self-attention layers to aggregate span-specific contextual representations. The resulting span embeddings are then fed into a linear classifier for final 
applicability condition prediction.
\item 
\textbf{Standard LoRA} on widely used LLMs, including Gemma2-9B \citep{team2024gemma}, Qwen2.5-7B \citep{qwen2025qwen25technicalreport}, Qwen3-4B \citep{yang2025qwen3technicalreport}, Gemma3-4B \citep{gemmateam2025gemma3technicalreport}, and its medical domain-adapted counterpart MedGemma-4B \citep{sellergren2025medgemmatechnicalreport}. We apply LoRA to all linear modules in the backbone models. In our main experiments, the LoRA rank is set to $8$.
\item 
\textbf{2-Shot Prompting} on DeepSeek-R1-70B \citep{deepseekai2025deepseekr1incentivizingreasoningcapability}, Llama3.3-70B \citep{grattafiori2024llama3herdmodels}, and Qwen2.5-72B \citep{qwen2025qwen25technicalreport}.
These models are selected as representative large-scale LLMs that are publicly accessible and have demonstrated strong performance on general reasoning and instruction-following tasks.
To construct 2-shot prompts, we randomly sample two drug-disease pairs from the training set. 
The prompt is presented in the Appendix~\ref{sec:prompt}
\end{itemize}

We adopt a linear learning rate schedule with warmup and decay following \cite{devlin-etal-2019-bert} when training SpanMarker models as well as LoRA models.
All experiments were conducted on a single NVIDIA H100 GPU.

\subsection{Implementation of Proposed Method}
For a fair comparison, our Role-Conditioned LoRA was implemented on the same backbone models as the standard LoRA baselines. We also use the same LoRA rank, setting it to $8$ in all main experiments. 
\citet{he-etal-2025-tablelora} empirically showed that applying task-specific LoRA adaptations
to key and value projections is an effective design choice. 
Following this, we apply our relation-role LoRA to the key and value projection layers, and employ standard LoRA to the remaining linear layers.

\subsection{Evaluation Metrics}
We evaluated the performance of the models at (a)~the span level, where only applicability condition spans are counted, and (b)~span and type, where both of the condition span and condition type matter. 
We employ Hard and Soft matching for evaluating condition spans. 

The final evaluation score is the average of F1 scores computed per sample. 
When an instance has no ground-truth applicability condition, the F1 is defined as $1.0$ if the model correctly predicts zero conditions, and $0.0$ otherwise.

\paragraph{Hard Matching} Under the hard matching criterion, a predicted span is considered correct only if it exactly matches the gold reference.

\paragraph{Soft Matching} We adopt the soft matching algorithm proposed by \citet{han2024empiricalstudyinformationextraction} to allow flexibility in evaluation. 
Soft matching relaxes the strict boundary requirement by considering both span containment and textual similarity between a predicted span and its gold reference. Specifically, a predicted span is regarded as a soft match if there exists a containment relation between the predicted and the gold reference span and their textual similarity exceeds a predefined threshold. 
In our experiments, the threshold is set to $0.5$.

\begin{table*}[t]
\centering
\small
\begin{tabular}{l c c c c c}
\hline
\multirow{2}{*}{\textbf{Model}} 
& \multirow{2}{*}{\makecell{\textbf{Trainable}\\\textbf{Params}}} 
& \multicolumn{2}{c}{\textbf{Span}} 
& \multicolumn{2}{c}{\textbf{Span and Type}} \\
\cline{3-6}
& 
& \textbf{Hard} & \textbf{Soft} 
& \textbf{Hard} & \textbf{Soft} \\
\hline

\multicolumn{6}{c}{\textit{SpanMarker}} \\
\hline
RoBERTa-base & $124.67\,\mathrm{M}$ & $34.86 \pm 2.12$ & $48.00 \pm 1.14$ & $34.81 \pm 2.19$ & $47.45 \pm 1.45$ \\
RoBERTa-large & $355.39\,\mathrm{M}$ & $37.81 \pm 2.11$ & $52.16 \pm 1.50$ & $37.81 \pm 2.11$ & $51.90 \pm 1.51$ \\
BERT-base-cased & $108.33\,\mathrm{M}$ & $30.73 \pm 2.19$ & $45.39 \pm 3.15$ & $30.69 \pm 2.23$ & $44.93 \pm 3.19$ \\
BERT-base-uncased & $109.51\,\mathrm{M}$ & $33.07 \pm 0.52$ & $45.20 \pm 1.64$ & $32.96 \pm 0.32$ & $44.80 \pm 1.70$ \\
BiomedBERT-base & $108.26\,\mathrm{M}$ & $39.35 \pm 1.06$ & $53.23 \pm 1.68$ & $39.35 \pm 1.06$ & $52.91 \pm 1.82$ \\
BioBERT-base-cased & $108.33\,\mathrm{M}$ & $33.57 \pm 1.25$ & $45.49 \pm 0.78$ & $33.38 \pm 1.56$ & $45.08 \pm 0.61$ \\
Bio\_ClinicalBERT & $108.33\,\mathrm{M}$ & $32.96 \pm 2.97$ & $46.40 \pm 1.90$ & $32.94 \pm 3.01$ & $46.03 \pm 2.01$ \\
\hline

\multicolumn{6}{c}{\textit{LLMs with LoRA Fine-tuning}} \\
\hline

\hline
Gemma2-9B$_{\text{LoRA}}$ & $27.01\,\mathrm{M}$ & $48.03 \pm 2.77$ & $59.09 \pm 2.38$ & $47.60 \pm 2.68$ & $58.30 \pm 1.99$ \\
Gemma2-9B$_{\text{RCLoRA}}$ & $28.39\,\mathrm{M}$ & \underline{$49.89 \pm 3.39$} & \underline{$59.57 \pm 3.01$} & \underline{$49.57 \pm 3.46$} & \underline{$58.83 \pm 3.02$} \\
Qwen2.5-7B$_{\text{LoRA}}$ & $20.19\,\mathrm{M}$ & $46.53 \pm 2.43$ & $56.12 \pm 1.75$ & $46.42 \pm 2.47$ & $55.51 \pm 1.69$ \\
Qwen2.5-7B$_{\text{RCLoRA}}$ & $20.42\,\mathrm{M}$ & \underline{$46.82 \pm 1.53$} & \underline{$56.99 \pm 0.65$} & \underline{$46.52 \pm 1.61$} & \underline{$56.23 \pm 0.56$} \\
Qwen3-4B$_{\text{LoRA}}$ & $16.52\,\mathrm{M}$ & $49.37 \pm 2.99$ & $59.84 \pm 3.12$ & $49.19 \pm 3.11$ & $59.18 \pm 3.30$ \\
Qwen3-4B$_{\text{RCLoRA}}$ & $17.11\,\mathrm{M}$ & \underline{$51.18 \pm 0.46$} & \underline{$\mathbf{60.62 \pm 0.61}$} & \underline{$50.91 \pm 0.45$} & \underline{$\mathbf{59.78 \pm 0.56}$} \\
Gemma3-4B$_{\text{LoRA}}$ & $14.90\,\mathrm{M}$ & $49.17 \pm 1.75$ & $56.53 \pm 0.99$ & $49.01 \pm 1.89$ & $56.10 \pm 1.19$ \\
Gemma3-4B$_{\text{RCLoRA}}$ & $15.46\,\mathrm{M}$ & \underline{$\mathbf{51.43 \pm 4.23}$} & \underline{$59.74 \pm 3.74$} & \underline{$\mathbf{51.20 \pm 4.17}$} & \underline{$59.11 \pm 3.59$} \\
MedGemma-4B$_{\text{LoRA}}$ & $14.90\,\mathrm{M}$ & $46.62 \pm 1.12$ & $55.35 \pm 1.02$ & $46.39 \pm 1.22$ & $54.51 \pm 0.90$ \\
MedGemma-4B$_{\text{RCLoRA}}$ & $15.46\,\mathrm{M}$ & \underline{$48.63 \pm 3.01$} & \underline{$57.40 \pm 2.60$} & \underline{$48.40 \pm 3.11$} & \underline{$56.56 \pm 2.63$} \\
\hline

\multicolumn{6}{c}{\textit{LLMs with 2-Shot Prompting}} \\
\hline

DeepSeek-R1-70B & - & $13.91 \pm 1.83$ & $35.31 \pm 2.42$ & $13.73 \pm 1.80$ & $35.00 \pm 2.46$ \\
Llama3.3-70B & - & $23.75 \pm 2.70$ & $32.37 \pm 0.73$ & $23.71 \pm 2.75$ & $32.12 \pm 0.77$ \\
Qwen2.5-72B & - & $23.04 \pm 1.44$ & $31.35 \pm 0.61$ & $22.82 \pm 1.37$ & $30.92 \pm 0.63$ \\
\hline
\end{tabular}
\caption{Performance comparison under Hard and Soft matching of span-only and span and type prediction: the best scores are indicated by \textbf{bold} fonts, and the \underline{underlines} indicate that the proposed method (\textbf{RCLoRA}) outperforms corresponding standard LoRA counterparts.}
\label{tab:main-results}
\end{table*}

\begin{table}[t]
\centering
\small
\begin{tabular}{l c c c c}
\hline
\multirow{2}{*}{\textbf{Method}} 
& \multicolumn{2}{c}{\textbf{Span}} 
& \multicolumn{2}{c}{\textbf{Span \& Type}} \\
\cline{2-5}
& \textbf{Hard} & \textbf{Soft} 
& \textbf{Hard} & \textbf{Soft} \\
\hline
LoRA (Avg.)      & $47.94$ & $57.39$   & $47.72$ & $56.72$ \\
RCLoRA (Avg.)    & $49.59$ & $58.86$   & $49.32$ & $58.10$ \\
$\Delta$        & $\mathbf{+1.65}$ & $\mathbf{+1.47}$   & $\mathbf{+1.60}$ & $\mathbf{+1.38}$ \\
$p$-value    & $0.013$ & $0.018$   & $0.015$ & $0.022$ \\
\hline
\end{tabular}
\caption{Statistical significance test results of RCLoRA across $5$ backbones on LLMs with LoRA fine-tuning ($3$ seeds x $5$ models = $30$ independent runs). Note: $p$-values are calculated using a paired $t$-test across all 15 experimental pairs.}
\label{tab:t-test}
\end{table}

\section{Results and Analysis}

\subsection{Main Results}

We report the average F1 scores with standard deviation across three random seeds for SpanMarker and LoRA fine-tuning, and five seeds for 2-shot prompting.
Table~\ref{tab:main-results} reports the main results under both Hard and Soft matching
at the span-only and span and condition type predictions across different methods and base LLMs. 

The proposed method achieved the best soft and hard F1 scores on Qwen3-4B and Gemma3-4B, respectively.
Table~\ref{tab:t-test} presents the statistical significance analysis of RCLoRA across five different LLM backbones. Notably, the proposed method significantly outperforms the standard LoRA fine-tuning ($p < 0.05$) consistently across all primary metrics.

\paragraph{SpanMarker.}
Among SpanMarker baselines, models initialized with domain-specific pretraining
generally achieve stronger performance.
In particular, BiomedBERT-base performs strongly despite its smaller model size for both span-only and span and type predicions.
This confirms the effectiveness of BiomedBERT’s extensive pretraining on abstracts from PubMed.\footnote{Note that the PubMed abstracts used in Drug-ACE are newer than those for BiomedBERT, thus there should be no data leakage.}

\paragraph{LLMs with LoRA fine-tuning.}
Different from SpanMarker models, MedGemma-4B was inferior to its general-domain counterpart, 
Gemma3-4B. 
This suggests that domain adaptation alone is not always sufficient for LLMs.

\paragraph{Few-shot prompting.}
Despite significantly larger model sizes, 2-shot prompting exhibits substantially
lower performance compared with fine-tuned models,
demonstrating the difficulty of the task and the necessity of supervised adaptation.
These results further indicate that prompt-based inference alone
is insufficient to conduct complex reasoning required by applicability condition extraction.

\subsection{Performance per Condition Types}
\begin{figure}[t]
  \centering
  \includegraphics[width=\linewidth]{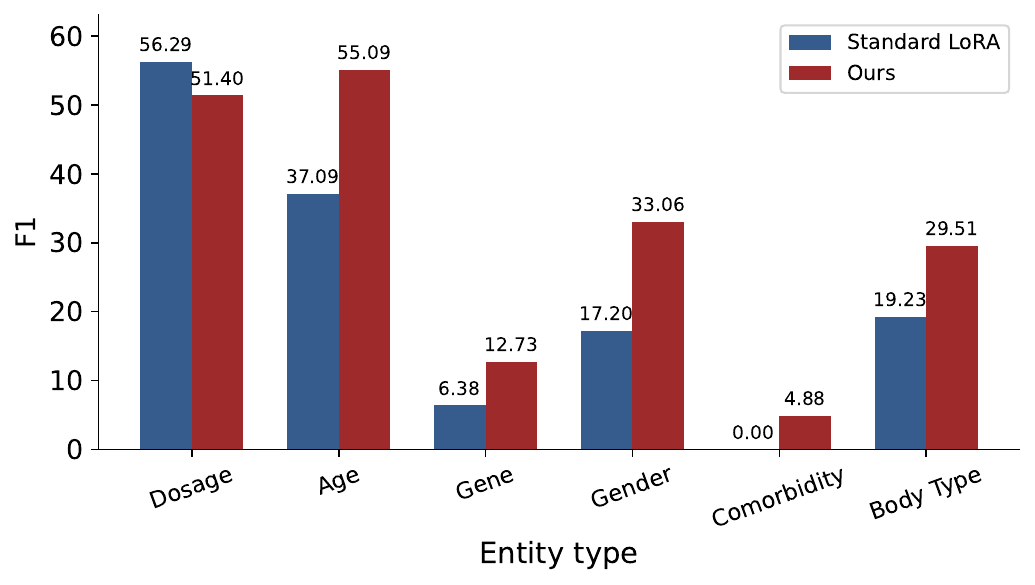}
  \caption{Performance comparison between standard LoRA and our method across different condition types}
  \label{fig:performance-type}
\end{figure}
Figure~\ref{fig:performance-type} illustrates the performance comparison between standard LoRA and our method across different condition types on Gemma2-9B. Our method outperforms the standard LoRA on all entity types: Age, Body Type, Comorbidity, Gender, and Gene, except Dosage. 
While the standard LoRA performs well on the most common type, i.e., Dosage, it struggles to identify other types. 
In particular, it yielded near-zero performance on Comorbidity-related conditions, whereas our method is able to capture meaningful signals for this type. 
We conjecture that condition types such as Gene and Comorbidity are challenging not only for their sparsity but also the highly specialized natures. 
Improvements on such types constitutes our future work. 

\subsection{Ablation Study}
The previous section demonstrated the effectiveness of encoding relation roles into LoRA. 
To further understand the effectiveness of our design, we compare to the following methods.

\paragraph{Marker:} 
For each drug-disease pair, we mark all occurrences of the corresponding entities in the input text using special tokens: \texttt{[Drug] entity [/Drug]} and \texttt{[Disease] entity [/Disease]}. We couple this marked input with a standard LoRA setup, keeping the model architecture untouched to isolate the effect of input-level markers. 

\paragraph{Role-Specific Vectors:} 
We directly add role-specific vectors at each layer instead of using low-rank matrices:
\begin{equation}
 {W}_{0} \mathbf{x} + B A \mathbf{x} + \mathbf{e}_{y}.
\end{equation}

\paragraph{Single-B Matrix:} 
We replace the two-matrix formulation of Eq.~(\ref{eq:rc lora}) with a single matrix, sharing the same $B$ matrix with LoRA:
\begin{equation}
 {W}_{0} \mathbf{x} + B A \mathbf{x} + B \mathbf{e}_{y}.
\end{equation}
This setting removes the separation between the transformation applied to the input vector
and that applied to the relation role embedding.

\paragraph{Random Roles:}
We input random ids of $0$, $1$, and $2$ to the relation embeddings, to distinguish the effect of using roles from simple increase in parameter sizes.

\paragraph{Results}
We employed Qwen3-4B as the backbone LLM for the ablation study for computational efficiency. 
For each setting, scores were averaged over three runs with different random seeds.
Table~\ref{tab:ablation} presents the results.
Compared with the standard LoRA, all variants lead to noticeable performance drops. 
In particular, the lower performance of the Marker setting suggests that explicitly highlighting drug and disease entities might inevitably cause the LLM to over-focus on the marked entities. This potentially leads the model to overlook critical contextual expressions in the biomedical context, which are significant for identifying the applicability conditions.
Furthermore, marking every occurrence of these entities throughout the text may introduce redundant signals that confuse LLMs' reasoning.
These results indicate that relation role requires distinctive LoRA matrix, and relation roles are crucial for the performance gain of the proposed method.

\begin{table}[t]
\centering
\small
\begin{tabular}{l c c c c}
\hline
\multirow{2}{*}{\textbf{Method}} 
& \multicolumn{2}{c}{\textbf{Span}} 
& \multicolumn{2}{c}{\textbf{Span \& Type}} \\
\cline{2-5}
& \textbf{Hard} & \textbf{Soft} 
& \textbf{Hard} & \textbf{Soft} \\
\hline
Standard LoRA      & $49.37$ & $59.84$   & $49.19$ & $59.18$ \\
Marker      & $49.26$ & $58.37$   & $48.99$ & $57.53$ \\
Role-Specific Vectors      & $43.93$ & $52.13$   & $43.72$ & $51.58$\\
Single B-Matrix    & $49.10$ & $58.57$   & $48.59$ & $57.64$ \\
Random Role        & $47.43$ & $56.85$   & $47.16$ & $56.27$ \\
RCLoRA (Proposed)    & $\mathbf{51.18}$ & $\mathbf{60.62}$ & $\mathbf{50.91}$ & $\mathbf{59.78}$ \\
\hline
\end{tabular}
\caption{Ablation study results on Qwen3-4B}
\label{tab:ablation}
\end{table}

\subsection{Effects of LoRA Ranks}
\begin{figure}[t]
  \centering
  \includegraphics[width=\linewidth]{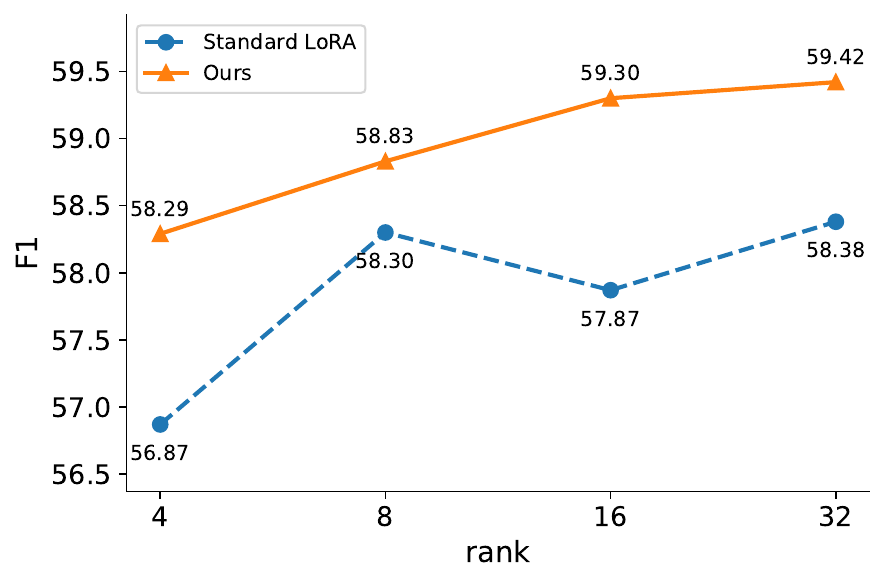}
  \caption{Effects of LoRA ranks on the Gemma2-9B}
  \label{fig:rank}
\end{figure}

Figure~\ref{fig:rank} shows the effects of LoRA ranks when using Gemma2-9B as the backbone LLM. 
For each rank, Soft F1 scores of span and condition type prediction were averaged over three runs with different random seeds.
Across different rank settings, our method consistently outperforms standard LoRA. Furthermore, the performance of our method enhance with larger rank, which aligns with the findings of \citet{zhangadaptive}. 

Table~\ref{tab:param-comparison} reports the corresponding number of trainable parameters under different rank settings for both standard LoRA and our method.
While our method slightly increases the number of trainable parameters at the same rank, the gaps are modest and scale consistently with the rank. 

\subsection{Effects of Soft Matching Threshold}
To investigate the impact of the similarity threshold used in soft matching, we conducted a sensitivity analysis comparing our primary model, Gemma3-4B-RCLoRA, against the standard Gemma3-4B-LoRA baseline on soft F1 scores of span and condition type. As shown in Table~\ref{tab:sensitivity}, while performance predictably decreases as the matching criterion becomes more stringent, RCLoRA consistently outperforms standard LoRA across all thresholds.\footnote{Increasing the similarity threshold makes the matching criterion more stringent, which reduces the number of True Positives. Given that the total counts of predictions and ground-truth are fixed, both Precision and Recall decrease monotonically.} We selected $0.5$ because hard matching ($1.0$) is overly restrictive for complex clinical spans. For example, if the target is "titrated to 50 mg once daily" and the model predicts "50 mg once daily" , exact matching would treat this as a complete failure. This would unfairly discard the model's success in capturing the core dosage information. A threshold of $0.5$ provides a reasonable trade-off that recognizes such clinically useful partial extractions.

\begin{table}[t]
\centering
\small
\begin{tabular}{l c c c c}
\hline
\multirow{2}{*}{\textbf{Method}} 
& \multicolumn{4}{c}{\textbf{Trainable Parameters}} \\
\cline{2-5}
& $\mathbf{r{=}4}$ & $\mathbf{r{=}8}$ & $\mathbf{r{=}16}$ & $\mathbf{r{=}32}$ \\
\hline
Standard LoRA & $13.5\,\mathrm{M}$ & $27.0\,\mathrm{M}$ & $54.0\,\mathrm{M}$ & $108.0\,\mathrm{M}$ \\
Ours          & $14.2\,\mathrm{M}$ & $28.4\,\mathrm{M}$ & $56.8\,\mathrm{M}$ & $113.5\,\mathrm{M}$ \\
\hline
\end{tabular}
\caption{Comparison of trainable parameters under different LoRA ranks on Gemma-9B}
\label{tab:param-comparison}
\end{table}

\begin{table}[t]
\centering
\small
\begin{tabular}{l c c c c c}
\hline
\multirow{2}{*}{\textbf{Method}} 
& \multicolumn{5}{c}{\textbf{Similarity Threshold}} \\
\cline{2-6}
& $\mathbf{0.1}$ & $\mathbf{0.3}$ & $\mathbf{0.5}$ & $\mathbf{0.7}$ & $\mathbf{0.9}$\\
\hline
LoRA & $64.27$ & $61.67$ & $57.45$ & $55.07$ & $51.39$ \\
RCLoRA & $\mathbf{68.88}$ & $\mathbf{67.26}$ & $\mathbf{63.16}$ & $\mathbf{60.49}$ & $\mathbf{56.26}$ \\
\hline
\end{tabular}
\caption{Performance comparison between LoRA and RCLoRA across different similarity thresholds}
\label{tab:sensitivity}
\end{table}

\section{Conclusion}
In this study, we introduced a novel task of drug–disease applicability condition extraction. 
We created the annotation dataset, named Drug-ACE, and proposed a method for applicability condition extraction that enhances LoRA by explicitly encoding relation roles between drugs and diseases. 

Our benchmarking results of conventional biomedical relation extraction methods on Drug-ACE highlight the challenges posed by applicability condition extraction and demonstrate the effectiveness of our method.

Our future work includes expanding the scale of the dataset, which may further improve model robustness. 
Improvement on challenging condition types, such as Gene and Comorbidity is also crucial. 
In addition, while this study focuses on therapeutic drug--disease relations, it should be worthy to extend to cover other biomedical relations, such as gene--disease interactions.

\section*{Limitations}

Despite the contributions of this study, several limitations remain.
First, the overall size of the proposed dataset is modest.
Although the dataset is carefully annotated,
scaling up the data size could further improve model robustness and enable more comprehensive empirical analysis.

In  addition, this study focuses exclusively on therapeutic drug--disease relations.
While this scope allows for a focused investigation of applicability conditions, other biologically meaningful relations, such as gene--disease interactions,
also exhibit conditional applicability and warrant further exploration. Extending the task and dataset to cover a broader range of biomedical relations
is an important direction for future research.

\paragraph{Potential Risks}
Our Drug-ACE dataset is intended for research purposes only. 
The annotated conditions reflect text-level reporting in biomedical research, which may include exploratory or experimental findings. Therefore, Drug-ACE should be viewed as an auxiliary tool for literature synthesis rather than a source of clinically validated medical truth.
Inappropriate use without expert validation may lead to misconduct in clinical environments.

\paragraph{AI Assistant Use}
We used AI assistants for improving writing; they were used exclusively for enhancing readability and correcting grammar. 
They did not contribute to the scientific content of the manuscript.

\section*{Acknowledgment}
We thank the domain expert, Narumi Tokunaga, and our annotators, Leonardo Ken Okumura and Miko Oikawa, for their significant contributions in Drug-ACE creation. 
This work was supported by Cross-ministerial Strategic Innovation Promotion Program (SIP) on ``Integrated Health Care System'' Grant Number JPJ012425.

\bibliography{custom}
\clearpage
\appendix

\section{Annotation Guideline}
\label{sec:guideline}
\subsection{Introduction}
This task is an applicability condition extraction for therapeutic drug-disease relations. The goal of the task is to extract applicability conditions (i.e., under what conditions a certain drug can treat a certain disease) for the mentioned drug from the given biomedical research literature.
\subsection{Task Definition}
Given a biomedical research literature (including a title and abstract) and a therapeutic drug–disease relation, annotators are required to identify the applicability condition(s) mentioned in the text that are relevant to the given drug–disease pair and assign a predefined label. Each annotation instance contains only one drug–disease pair. However, a single biomedical document may include multiple therapeutic drug–disease relations. In such cases, please perform the annotation separately for each specific drug–disease pair provided.
\subsection{Inclusion and Exclusion Criteria}
The corpus may contain biomedical literature related to cell or animal experiments rather than human clinical studies. 
Annotators are required to verify whether the given biomedical literature pertains to a human clinical study. 
If the given literature is not about a human clinical study, the entire literature must be skipped.
Additionally, the provided drug--disease relation may not be therapeutic. If the relation is determined to be non-therapeutic, the specific annotation instance should be skipped.
\subsection{Condition Types}
We assign condition type labels to each extracted drug-disease condition. The condition types are as follows:
\begin{itemize}
  \item \textit{Dosage}, indicates the dosage or amount of the drug administered, including specific dosage values, ranges that are required for effective or safe treatment.
  \item \textit{Age}, specifies the patient age or age group, including explicit ages or age-related categories.
  \item \textit{Gene}, specifies genetic characteristics of patients, such as the presence of particular genes, that influence drug response or treatment suitability.
  \item \textit{Gender}, indicates the biological sex or gender of the target patient population,
  such as male or female, when treatment applicability differs across genders.
  \item \textit{Comorbidity}, refers to the presence of one or more additional pre-existing diseases, disorders, or risk factors other than the index disease under investigation, which may affect the drug’s applicability or therapeutic effect.
  \item \textit{Body Type} describes general body characteristics or physical conditions of patients, including pregnancy, obesity, underweight status, or other body composition–related factors that may influence treatment outcomes.
\end{itemize}
\subsection{Annotation Scope}
\begin{itemize}
  \item An applicable condition is defined as any contextual factor (e.g., patient attribute) that is explicitly stated or strongly implied to influence the effectiveness or safety of the drug--disease relation. 
  
   (e.g., "...... Long-term oral warfarin is recommended in \colorbox{green}{pediatric} Kawasaki disease patients with \colorbox{purple!30}{large coronary artery aneurysms}. ...... \textbf{Drug--Disease Pair}: warfarin-Kawasaki disease" \textbf{Annotated Conditions}: (1) \textit{``pediatric''} $\rightarrow$ Age (2) \textit{``large coronary artery aneurysms''} $\rightarrow$ Comorbidity).
  \item Not every given biomedical research literature contains applicability conditions for the given therapeutic drug -- disease relation. 

  (e.g., "...... Moreover, lisinopril and nifedipine appear to be capable of reducing bcl-2 concentrations, with potentially beneficial effects on vascular modifications in patients with hypertension. ...... \textbf{Drug--Disease Pair}: nifedipine - hypertension" \textbf{Annotated Conditions}: No applicability condition.
  
  \item When the applicability conditions present, annotators are instructed to extract them even if they are mentioned as general factors without specific values. 

  (e.g., "...... Patient \colorbox{blue!30}{height} is the main factor determining warfarin dosage, while genotype effects on warfarin dosage vary among studies. ...... \textbf{Drug--Disease Pair}: warfarin-Kawasaki disease" \textbf{Annotated Conditions}: \textit{``height''} $\rightarrow$ Body Type).
  
  \item Applicability conditions described quantitatively, such as demographic characteristics presented with percentages, are also considered within the scope of valid extractions.

  (e.g., "...... PATIENTS: These patients were mostly \colorbox{orange!50}{men (57\%)} \colorbox{green}{older than 30 years (56\%)} with pulmonary obstruction, ...... \textbf{Drug--Disease Pair}: zafirlukast - asthma" \textbf{Annotated Conditions}: (1) \textit{``men (57\%)''} $\rightarrow$ Gender (2) \textit{``older than 30 years (56\%)''} $\rightarrow$ Age).
\end{itemize}

\section{Prompt for 2-shot Prompting}
\label{sec:prompt}

The example~\ref{ex:1} shows the prompt template we use for extracting applicability conditions  in 2-shot prompting.

\newtheorem{example}{Example}[section] 
\begin{example} \label{ex:1}
You are a skilled biomedical text annotator. Given the title, abstract, and a therapeutic drug–disease relation, extract all applicability condition spans mentioned in the text that mention under what conditions the drug is used to treat the disease.

    Applicability conditions include:
    
    - Dosage: indicates the dosage or amount of the drug administered, including specific dosage values, ranges that are required for effective or safe treatment.
    
    - Age: specifies the patient age or age group, including explicit ages or age-related categories.
    
    - Gender: indicates the biological sex or gender of the target patient population,
    
    - Comorbidity: refers to the presence of one or more additional pre-existing diseases, disorders, or risk factors other than the index disease under investigation, which may affect the drug’s applicability or therapeutic effect.
    
    - Body type: describes general body characteristics or physical conditions of patients, including pregnancy, obesity, underweight status, or other body composition–related factors that may influence treatment outcomes.
    
    - Gene: specifies genetic characteristics of patients, such as the presence of particular genes, that influence drug response or treatment suitability.
    
    For each identified span, you need to specify the corresponding label, please return it in the format:
    "['Span: <span> | Label: <type>', 'Span: <span> | Label: <type>', ...]"

    Please strictly follow the format I gave you.
    Only include spans that are explicitly mentioned in the context.
    Do not infer conditions beyond what is supported by the text.

    If no applicability condition is mentioned, return an empty list: []

    An example:

    TITLE: \{Example 1 title\}
    
    ABSTRACT: \{Example 1 abstract\}
    
    DRUG - DISEASE RELATION: \{Example 1 drug\} -- \{Example 1 disease\}
    
    APPLICABILITY CONDITIONS:
    
    \{List of span -- type pairs\}

    Another example:

    TITLE: \{Example 2 title\}
    
    ABSTRACT: \{Example 2 abstract\}
    
    DRUG - DISEASE RELATION: \{Example 2 drug\} -- \{Example 2 disease\}
    
    APPLICABILITY CONDITIONS:
    
    \{List of span -- type pairs\}

    Now, your turn:

    TITLE: \{Input title\}
    
    ABSTRACT: \{Input abstract\}
    
    DRUG - DISEASE RELATION: \{Input drug\} -- \{Input disease\}
    
    APPLICABILITY CONDITIONS:
\end{example}

\end{document}